\documentclass{article}

\usepackage[english]{babel}
\usepackage{graphicx}
\usepackage{caption}
\usepackage{subcaption}
\usepackage{float} 
\usepackage{url}
\usepackage[colorlinks=true, allcolors=blue]{hyperref} 

\usepackage[letterpaper,top=2cm,bottom=2cm,left=3cm,right=3cm,marginparwidth=1.75cm]{geometry}

\usepackage{amsmath}

\title{A Predictive and Optimization Approach for Enhanced Urban Mobility Using Spatiotemporal Data}
\author{
    \begin{minipage}[t]{0.45\textwidth}
        \centering
        Shambhavi Mishra\thanks{shambhavimishra328@gmail.com}\\
        Student, Department of Information Technology
    \end{minipage}%
    \hfill
    \begin{minipage}[t]{0.45\textwidth}
        \centering
        Dr. T. Satyanarayana Murthy\thanks{tsmurthy\_it@cbit.ac.in}\\
        Associate Professor, Department of Information Technology
    \end{minipage}
}
\date{}

\begin{document}
\maketitle

\begin{abstract}
In modern urban centers, effective transportation management poses a significant challenge, with traffic jams and inconsistent travel durations greatly affecting commuters and logistics operations. This study introduces a novel method for enhancing urban mobility by combining machine learning algorithms with live traffic information. We developed predictive models for journey time and congestion analysis using data from New York City's yellow taxi trips. The research employed a spatiotemporal analysis framework to identify traffic trends and implemented real-time route optimization using the GraphHopper API. This system determines the most efficient paths based on current conditions, adapting to changes in traffic flow. The methodology utilizes Spark MLlib for predictive modeling and Spark Streaming for processing data in real-time. By integrating historical data analysis with current traffic inputs, our system shows notable enhancements in both travel time forecasts and route optimization, demonstrating its potential for widespread application in major urban areas. This research contributes to ongoing efforts aimed at reducing urban congestion and improving transportation efficiency through advanced data-driven methods.

\end{abstract}
\vspace{0.5cm} 


\section{Introduction}

The ability of people and goods to move within urban areas, known as urban mobility, has become a vital aspect of modern city planning and growth. The rise of large metropolitan areas, especially in the 21st century, has created a greater need for efficient, sustainable, and flexible transportation networks. As more people move to cities worldwide, there is an increasing urgency to tackle mobility issues such as traffic jams, environmental harm, and inefficient use of resources. Well-functioning urban mobility systems are essential for any city, directly affecting economic output, living standards, and environmental sustainability.

Traffic congestion, in particular, is a major problem in cities across the globe. It leads to delays, higher fuel use, and increased emissions, resulting in economic losses and negative environmental effects. As global populations grow and more people move to urban areas, cities struggle to manage limited road space while accommodating more vehicles. In many cases, existing infrastructure has not kept up with the rapid increase in demand, causing significant inefficiencies in urban transportation systems. Studies show that urban congestion not only hinders economic productivity but also negatively impacts the mental and physical well-being of city dwellers.

Various strategies have been employed over time to reduce congestion, from expanding roads and building new infrastructure to implementing traffic management solutions. However, traditional methods have often fallen short in addressing the root cause due to the ever-changing and complex nature of urban traffic patterns. The incorporation of data-driven solutions, made possible by advancements in real-time traffic monitoring, data analysis, and artificial intelligence, offers a promising new approach to tackling urban mobility challenges. By harnessing big data, predictive modeling, and optimization algorithms, cities can more effectively control traffic flows, decrease congestion, and improve overall mobility.

Recent research in urban mobility has increasingly emphasized the use of machine learning, traffic optimization models, and real-time data to forecast traffic patterns and recommend optimal routes for commuters. Numerous studies have investigated the application of geographic information systems (GIS), spatiotemporal analysis, and various algorithms to model traffic conditions. Furthermore, the development of intelligent transportation systems (ITS) and smart city initiatives has gained traction in recent years, with researchers exploring ways to integrate traffic data, public transportation, and infrastructure to create a more efficient urban mobility framework. Despite these advancements, existing research has often neglected the potential of combining real-time traffic data with route optimization models for comprehensive traffic management. Identifying a Research Gap
Despite advancements in urban mobility understanding and optimization, a significant void exists in the effective combination of real-time traffic information with optimization models for instant route recommendations. Most current research emphasizes predictive models that are either fixed or rely on historical data, thus reducing their precision in fluctuating traffic situations. Moreover, a substantial portion of studies neglects the opportunity to merge spatiotemporal traffic examination with immediate optimization tools like GraphHopper to offer on-the-spot solutions for travelers. This gap is especially evident in large, ever-changing urban settings where traffic conditions shift rapidly and unexpectedly.

The primary research inquiry arising from this gap is: In what way can live traffic data be efficiently incorporated with machine learning and optimization algorithms to deliver dynamic, precise route suggestions in urban environments? Tackling this question is vital for enhancing urban mobility systems by alleviating traffic congestion and boosting transportation network efficiency.

This study's research fills in a number of holes in the field of optimizing urban transportation by utilizing real-time data integration and cutting-edge technologies. This study fills the gap in existing systems' inability to seamlessly incorporate real-time traffic data for instantaneous decision-making by integrating APIs such as HERE, OSRM, and GraphHopper, offering instantaneous route choices based on traffic circumstances at the moment. Moreover, whereas a lot of traditional research focuses on real-time optimization or historical data analysis, very few successfully integrate the two to foresee congestion. In order to get around that restriction, this study uses machine learning models created with Spark MLlib. These models estimate congestion and improve route selection dynamically by utilizing both historical and real-time data. Spatiotemporal analysis integration provides a sophisticated grasp of how traffic patterns vary according to the day of the week or the time of day, an aspect that static models frequently omit. Through the analysis of these variances, the study adds fresh understanding of how urban congestion changes over time while also increasing route efficiency. This all-encompassing strategy makes sure that the system not only responds to present circumstances but also foresees and adjusts to future patterns of congestion, providing a more proactive and astute response to issues with urban mobility.

The innovation of this research lies in its comprehensive approach to urban mobility. Unlike existing studies that focus solely on either real-time traffic data or predictive modeling, this investigation combines both to deliver a more accurate and dynamic solution for optimizing urban traffic flows. By utilizing a combination of traffic data, machine learning models, and optimization algorithms, the study presents a thorough solution to urban mobility challenges, particularly in highly congested environments.

\section{Related Work}

\subsection{Transportation System Transformation Using Big Data}

Big data is driving a major shift in the transportation sector that will improve decision-making, increase operational efficiency, and support sustainable practices. Big data analytics integration makes it easier for transportation authorities to monitor traffic patterns and comprehend how they affect air quality by enabling them to make well-informed judgments based on both historical and current data \cite{bachechi2021big}. Authorities and individuals can see traffic flow and react quickly to accidents or congestion thanks to real-time monitoring systems like the Trafair Traffic Dashboard, which use big data to deliver live traffic updates \cite{bachechi2021big}.
Furthermore, transportation systems can optimize traffic flow and reduce congestion by analyzing spatio-temporal data. Understanding how different factors influence traffic dynamics requires the use of traffic simulations and sensor data \cite{bachechi2021big}. In order to inform policies targeted at lowering emissions and promoting cleaner transportation options, big data analytics is also essential for assessing how transportation decisions affect air pollution . Big data is an essential part of smart city initiatives since it unifies many urban data sources, improving urban administration and planning. Moreover, proactive approaches to peak traffic scheduling and pollution mitigation are made possible by predictive analytics that is based on historical data trends .
Nonetheless, the intricacies involved in merging many data sources and the fluctuations in data caliber are frequently disregarded in current research, potentially impacting the validity of the conclusions drawn. Furthermore, although a lot of research highlights the advantages of big data in urban transportation, less attention is paid to the restrictions on data privacy and the moral issues surrounding data collection.

\subsection{Machine Learning and Artificial Intelligence for Mobility Optimization}

Machine learning (ML) and artificial intelligence (AI) are now crucial tools for improving mobility optimization and forecasts. Due to their efficiency in handling complicated data patterns, deep learning techniques like Convolutional Neural Networks (CNN) and Long Short-Term Memory (LSTM) models are frequently used for traffic flow prediction \cite{ushakov2022big}. Furthermore, traffic signal optimization can be achieved by utilizing data from connected vehicles, which will enhance traffic flow and lessen congestion overall \cite{ulvi2024urban}. Using sensor data, Multi-Agent Systems (MAS) provide very accurate traffic condition forecasts and effective urban traffic control \cite{ulvi2024urban}.
Furthermore, the management of urban transit systems requires the application of sophisticated data mining techniques, especially when modeling temporal relationships in time-series data for passenger flow in urban rail transit. Federated learning techniques are employed in deep learning applications to improve performance in ride-hailing service optimization and urban traffic congestion management \cite{ushakov2022big}. Even with these developments, research on the scalability of AI and ML systems is still lacking, especially given the increasing amount of data. Numerous previous research could not have sufficiently taken into account the potential biases in training data as well as how robust these models are under different circumstances

\subsection{Spatio-Temporal Analysis in Urban Mobility}

The utilization of spatiotemporal analysis has become an advantageous approach in the interpretation of complex patterns of urban transportation. Researchers can obtain a more thorough picture of urban mobility trends by merging several geographic data sources, such as mobile positioning data, smart card records, and GPS \cite{cho2021clustered}. In order to promote individualized services and efficient urban planning, mobility pattern analysis allows for the prediction of future traffic flows and destinations \cite{cho2021clustered}.

Moreover, knowing how people behave differently on the weekends and during the day improves the accuracy of mobility models and increases the effectiveness of the transportation system \cite{liu2021urban}. User recommendations can be made more precisely and uniquely by integrating personal data and preferences into spatiotemporal analysis . For sustainable city expansion, employment creation, and overall development, it is also imperative to analyze spatiotemporal patterns of urban mobility \cite{cho2021clustered}. Unfortunately, the problems of data heterogeneity and granularity are frequently ignored in existing publications, which can reduce the efficacy of these analyses. Further investigation is also necessary on the implications of data security and user privacy in spatiotemporal data use

\subsection{Dynamic Transportation Systems and Real-Time Route Optimization}

Improving urban functionality and lowering traffic congestion require real-time route optimization. Urban transportation systems, enhanced by real-time optimization, are essential to the functioning, expansion, and general well-being of a city \cite{ma2023overview}. Predictive models can anticipate congestion and suggest alternate routes by utilizing spatiotemporal traffic features \cite{ushakov2022big}. To evaluate urban travel patterns and create efficient traffic management plans, data-driven methods and mathematical models underpin data-informed traffic control .By analyzing enormous volumes of data from linked cars and mobile services, the combination of Intelligent Transportation Systems (ITS) with real-time route optimization might significantly improve urban mobility . According to \cite{ulvi2024urban}, Multi-Agent Systems make use of data from wireless sensor networks to improve parking space use, regulate road intersections, and forecast traffic flow. However, real-time data processing and the computing demands of these systems are typically ignored in current research, which can restrict their scalability and usefulness in larger urban contexts.

\subsection{Obstacles in Handling Large-Scale, High-Velocity Data}
Large-scale, high-velocity data management from various sources in real-time poses several obstacles even with notable breakthroughs. Consolidating data from several sources with different formats, structures, and update frequencies is a complicated process . Maintaining the quality of incoming data depends on ensuring data integrity and putting in place effective data cleaning procedures . Moreover, systems need to be able to handle expanding data quantities as urban areas grow and the number of sensors rises .

High-performance computer resources and effective algorithms are needed to meet the need for instantaneous processing . Significant issues also arise from the need to store vast amounts of data while guaranteeing prompt retrieval for analysis . Another ongoing difficulty for public authorities and citizens is the visualization and interaction with complicated data formats . Furthermore, seamless operations depend on various systems and technologies being compatible with one another . There are gaps in our knowledge on how to efficiently manage large-scale data in dynamic urban environments since current studies frequently do not adequately investigate these difficulties.
\subsection{Future Prospects}
Big data, real-time analytics, and technologies like Apache Spark have a great deal of potential to improve urban mobility in the long run. These technologies will become more and more important in creating efficient, sustainable, and livable urban settings as cities expand and change. Big data analytics will make it possible to monitor urban transportation systems more thoroughly and accurately. This will allow for the development of exact predictions and optimization strategies that have the potential to completely transform public transportation and traffic management. This is especially true when paired with IoT devices and sensors.

Instantaneous decision-making in traffic management will be made easier by real-time analytics driven by technologies such as Apache Spark. This will lead to personalized travel recommendations, adaptive public transportation routing, and dynamic traffic signal control—all of which are updated in real-time based on the current situation. Additionally, incorporating machine learning and artificial intelligence with these big data systems promises to unearth deeper insights into urban mobility patterns. This could result in improved urban planning, autonomous transportation systems that adapt to changing urban conditions, and predictive maintenance of transportation infrastructure.

Urban mobility has a bright future thanks to the combination of big data analytics, AI/ML methods, spatiotemporal analysis, and real-time optimization. Our understanding, management, and optimization of urban transportation systems are changing as a result of these technologies. Nonetheless, there are still difficulties, especially with data management and real-time processing. Future studies should concentrate on resolving these issues and utilizing these technologies to their fullest extent in order to develop more intelligent, effective, and sustainable urban transportation systems.

\section{Methodology}

The Urban Mobility Optimization project employs a multifaceted approach to comprehensively analyze urban traffic patterns. This strategy incorporates various stages, including data acquisition, refinement, predictive analysis, real-time traffic evaluation, spatiotemporal examination, and route enhancement. The following sections elucidate each component, offering a thorough understanding of the research methodologies utilized.

\begin{figure}[h!]
    \centering
    \includegraphics[width=0.6\textwidth]{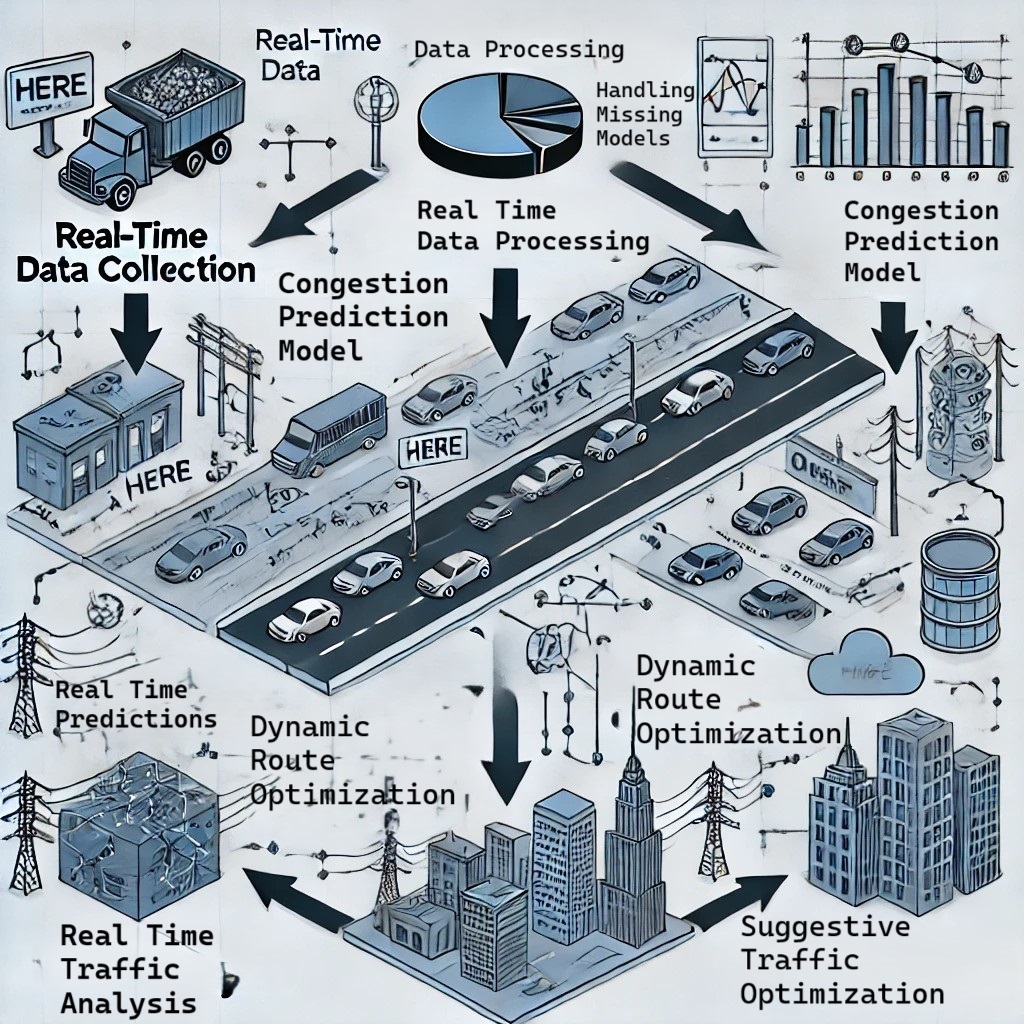}
    \caption{}
    \label{fig:image1}
\end{figure}

\subsection{Data Collection}
The foundation of this study rests on extensive data collection from diverse sources. Initially, the HERE API was utilized to amass comprehensive datasets encompassing historical and real-time traffic information, such as vehicle speeds, congestion intensity, and road status. This API also provided access to current traffic incidents that could substantially affect travel durations. Due to subsequent accessibility challenges, the project shifted to employing the Open Source Routing Machine (OSRM) and GraphHopper, which deliver high-efficiency routing capabilities and up-to-the-minute urban traffic data updates.
Furthermore, the data collection process incorporated New York City (NYC) taxi information, offering insights into traffic trends based on actual travel behaviors within a dynamic urban setting. The NYC yellow taxi trip data included attributes like pickup and drop-off points, time stamps, journey distances, and fare amounts. This information was crucial for understanding traffic flow dynamics, as it mirrored real-world vehicle interactions in the urban landscape.
This comprehensive data acquisition phase established the primary input for subsequent analyses, enabling a robust examination of traffic patterns and route optimization based on both current and historical conditions.

\subsection{Data Preprocessing}
Following data collection, an extensive refinement phase was undertaken to ensure data quality and suitability for analysis. This process involved several key steps:

\begin{description}
    \item[Data Cleaning] The raw information gathered from APIs and the NYC taxi dataset often contained anomalies, missing values, and inconsistencies. Data purification was performed using Python libraries such as Pandas to identify and rectify these issues. Techniques like interpolation were employed to fill in missing values, while outlier detection methods were utilized to filter out anomalous data points that could skew results.
    
    \item[Data Transformation] To prepare the information for analysis, transformations were applied to convert the dataset into a format compatible with machine learning models. This included converting categorical variables into numerical representations using one-hot encoding. Additionally, continuous variables were normalized to ensure they were on a similar scale, which is crucial for many machine learning algorithms.
    
    \item[Feature Engineering] New attributes were created to enhance the predictive power of the models. Temporal features (e.g., hour of the day, day of the week) were generated to capture daily traffic patterns. Historical traffic data was aggregated to produce features representing congestion levels over prior periods.
\end{description}

\subsection{Predictive Modeling}
In the Urban Mobility Optimization project, predictive modeling serves a crucial function by offering projections of future traffic patterns based on historical information. This approach involves the application of statistical algorithms to examine and anticipate traffic congestion, enabling proactive traffic management strategies. The study employed two primary algorithms: K-Means Clustering and Linear Regression, each chosen for its appropriateness to the specific analysis required.

\subsubsection{K-Means Clustering}
To categorize traffic data into distinct groups based on past congestion levels, K-Means Clustering was utilized. This algorithm excels at recognizing patterns in unlabeled data and is commonly applied in traffic analysis to classify various traffic conditions. The K-Means algorithm functions as follows:

\begin{enumerate}
    \item \textbf{Initialization}: The process begins with the selection of $K$ initial centroids, representing cluster centers. These can be randomly chosen from the dataset or determined through more advanced methods like K-Means++ for improved initialization.
    
    \item \textbf{Assignment Step}: Each data point is allocated to the nearest centroid, forming $K$ clusters. Typically, the distance is measured using Euclidean distance, calculated as:
    \begin{equation}
        d(x,\mu_i) = \sqrt{\sum_{j=1}^n (x_j - \mu_{ij})^2}
    \end{equation}
    Where:
    \begin{itemize}
        \item $d(x,\mu_i)$ is the distance between a data point $x$ and the centroid $\mu_i$.
        \item $x_j$ are the components of the data point, and $\mu_{ij}$ are the components of the centroid.
    \end{itemize}
    
    \item \textbf{Update Step}: After all data points are assigned to clusters, the centroids are recalculated by taking the mean of all points in each cluster. The new centroid is given by:
    \begin{equation}
        \mu_i = \frac{1}{n_i} \sum_{j=1}^{n_i} x_j^{(i)}
    \end{equation}
    Where:
    \begin{itemize}
        \item $n_i$ is the number of data points in cluster $i$.
        \item $x_j^{(i)}$ are the data points assigned to cluster $i$.
    \end{itemize}
    
    \item \textbf{Convergence}: The algorithm iterates through the assignment and update steps until the centroids no longer change significantly or a predefined number of iterations is reached. The objective function to be minimized is the total within-cluster variance, represented mathematically as:
    \begin{equation}
        J = \sum_{i=1}^K \sum_{j=1}^{n_i} \| x_j^{(i)} - \mu_i \|^2
    \end{equation}
    Where:
    \begin{itemize}
        \item $J$ is the total within-cluster variance.
        \item $K$ is the number of clusters.
        \item $x_j^{(i)}$ represents the data points in cluster $i$.
        \item $\mu_i$ is the centroid of cluster $i$.
    \end{itemize}
\end{enumerate}

By implementing K-Means Clustering, the research aimed to identify specific traffic patterns and periods of peak congestion. This clustering method facilitated the recognition of traffic conditions corresponding to varying congestion levels, allowing urban planners and traffic managers to comprehend overall traffic dynamics in urban areas.

\subsubsection{Linear Regression}
Linear Regression was employed to forecast future traffic congestion levels using historical data, thereby establishing a quantitative link between traffic conditions and influencing factors. This model is particularly effective for understanding how independent variables impact a dependent variable—in this case, how time-related elements affect traffic congestion. The Linear Regression model operates as follows:

\begin{enumerate}
    \item \textbf{Model Formulation}: The relationship between the dependent variable (traffic congestion levels) and independent variables (e.g., time of day, day of the week, historical traffic patterns) can be expressed mathematically as:
    \begin{equation}
        Y = \beta_0 + \beta_1X_1 + \beta_2X_2 + \ldots + \beta_nX_n + \epsilon
    \end{equation}
    Where:
    \begin{itemize}
        \item $Y$ is the dependent variable (congestion level).
        \item $\beta_0$ is the intercept of the regression line.
        \item $\beta_1, \beta_2, \ldots, \beta_n$ are the coefficients representing the influence of each independent variable.
        \item $X_1, X_2, \ldots, X_n$ are the independent variables (e.g., time of day, day of the week).
        \item $\epsilon$ is the error term, accounting for variability not explained by the model.
    \end{itemize}

    \item \textbf{Training the Model}: The model is trained using a preprocessed dataset, which involves:
    \begin{itemize}
        \item Dividing the dataset into training and testing subsets to evaluate the model's performance.
        \item Fitting the model to the training data using a method such as Ordinary Least Squares (OLS) to minimize the sum of the squared differences between observed and predicted values:
        \begin{equation}
            \min_{\beta_0, \beta_1, \ldots, \beta_n} \sum_{i=1}^m (Y_i - \hat{Y}_i)^2
        \end{equation}
        Where:
        \begin{itemize}
            \item $m$ is the number of observations.
            \item $Y_i$ is the actual congestion level.
            \item $\hat{Y}_i$ is the predicted congestion level.
        \end{itemize}
    \end{itemize}

    \item \textbf{Model Evaluation}: The predictive accuracy of the Linear Regression model is assessed using metrics such as Mean Absolute Error (MAE) and Root Mean Square Error (RMSE):
    \begin{itemize}
        \item Mean Absolute Error (MAE):
        \begin{equation}
            MAE = \frac{1}{m} \sum_{i=1}^m |Y_i - \hat{Y}_i|
        \end{equation}
        \item Root Mean Square Error (RMSE):
        \begin{equation}
            RMSE = \sqrt{\frac{1}{m} \sum_{i=1}^m (Y_i - \hat{Y}_i)^2}
        \end{equation}
    \end{itemize}
\end{enumerate}

By utilizing Linear Regression, the study aimed to accurately predict congestion levels by modeling the relationships between historical traffic data and various temporal factors. This predictive capability is essential for implementing proactive traffic management strategies, thus optimizing urban mobility.

\subsubsection{Real-Time Traffic Analysis}
A key component of the methodology is real-time traffic analysis, which enables immediate route adjustments based on current traffic conditions. This dynamic system uses real-time data from the Open Source Routing Machine (OSRM) and GraphHopper, allowing continuous monitoring of traffic dynamics. The analysis process consists of several crucial steps:

\begin{enumerate}
    \item \textbf{Data Retrieval}: Real-time traffic data is obtained at regular intervals using the OSRM and GraphHopper APIs. This information includes essential details such as current traffic speeds, estimated travel times, and reports of road incidents. The APIs provide a constant stream of updated traffic information, which is vital for assessing real-time conditions.

    \item \textbf{Comparison with Predictive Models}: The system compares current traffic situations with forecasts developed earlier in the project. This assessment aims to detect notable differences between anticipated congestion and actual conditions. A predetermined threshold determines when these variances necessitate route recommendation updates. The system initiates a reassessment of routing suggestions if real-time data shows traffic is considerably worse than expected, such as surpassing a specific percentage of predicted travel duration.

    \item \textbf{Dynamic Route Adjustment}: When substantial deviations are identified, the system employs GraphHopper's routing algorithm to recalculate optimal paths. GraphHopper utilizes either Dijkstra's or A* search algorithms to determine the shortest route based on current traffic information, which is essential for reducing delays and optimizing travel times.
    \begin{itemize}
        \item \textbf{Dijkstra's Algorithm} is a renowned method for identifying the shortest paths between graph nodes. The fundamental steps of this algorithm are:
        \begin{enumerate}
            \item Set all vertex distances to infinity, except the starting vertex, which is set to zero.
            \item Label all nodes as unvisited and create an unvisited set containing all nodes.
            \item For each unvisited node, examine neighbouring nodes and calculate their tentative distances through the current node.
            \item Update a neighbour's distance if the calculated distance is less than the previously recorded one.
            \item After examining all neighbours, mark the current node as visited and remove it from the unvisited set.
            \item Repeat this process until all nodes have been evaluated, resulting in the shortest path from the start node to all other nodes in the graph.
        \end{enumerate}
        Dijkstra's algorithm performs efficiently for graphs with a reasonable number of vertices and edges, making it appropriate for urban traffic networks.

        \item \textbf{A* Search Algorithm} is an alternative method used to enhance efficiency. It improves upon Dijkstra's algorithm by incorporating heuristics to estimate the cost of reaching the destination from each node. The A* algorithm formula combines the actual distance from the start node ($g(n)$) with a heuristic estimate ($h(n)$) of the distance to the goal:
        \begin{equation}
            f(n) = g(n) + h(n)
        \end{equation}
        Where $f(n)$ represents the total estimated cost of the most economical solution through node $n$, $g(n)$ is the cost from the start node to $n$, and $h(n)$ is the estimated cost from $n$ to the goal. The heuristic function $h(n)$ is often defined using Euclidean or Manhattan distance, depending on the urban area's grid structure.
    \end{itemize}
\end{enumerate}

This ongoing analysis improves the system's responsiveness, enabling prompt and efficient route adjustments as traffic conditions change.

\subsection{Spatiotemporal Analysis}
Spatiotemporal analysis enhances real-time traffic monitoring by offering insights into traffic trends across time and space. This approach examines traffic data to identify recurring congestion areas and comprehend overall traffic behavior. The process consists of several stages:

\begin{enumerate}
    \item \textbf{Temporal Analysis}: Traffic information was evaluated to determine peak congestion periods by consolidating data into specific time frames, such as hourly or daily intervals. This investigation uncovered crucial times when traffic congestion is most intense, which is vital for urban development and traffic control. Methods like time series analysis were utilized to examine how congestion fluctuates throughout the day and week, enabling targeted interventions during high-traffic periods.

    \item \textbf{Spatial Analysis}: Using geospatial information, congestion levels were plotted across various urban regions. This spatial analysis was conducted using tools such as Geopandas and Folium, allowing for the visualization of congestion hotspots on maps. Geospatial analysis techniques included:
    \begin{itemize}
        \item \textbf{Kernel Density Estimation (KDE)}: A statistical approach used to estimate the probability density function of a random variable. In this instance, KDE was applied to visualize the concentration of traffic incidents in particular areas, generating a heatmap representation of congestion levels.
        \item \textbf{Geographical Information Systems (GIS)}: GIS tools were employed to analyze spatial relationships between various urban features and traffic conditions. By combining different data types (e.g., road networks, traffic incidents, and historical congestion data), GIS provided a comprehensive view of urban traffic dynamics.
    \end{itemize}

    \item \textbf{Heatmap Generation}: Heatmaps were generated to visually depict congestion levels, offering an intuitive understanding of traffic dynamics. These heatmaps highlighted areas with frequent congestion, assisting urban planners and traffic managers in making informed decisions about infrastructure improvements and traffic management strategies. The visual representation of data helped identify critical locations where traffic flow could be enhanced.
\end{enumerate}

Through spatiotemporal analysis, valuable insights into traffic patterns were obtained, which are essential for understanding the underlying causes of congestion and developing long-term mitigation strategies.

\subsection{Route Optimization}
The culmination of the methodology is the route optimization process, which utilizes the collected and analyzed data to dynamically optimize routes for users based on real-time traffic conditions. The route optimization workflow involves the following steps:

\begin{enumerate}
    \item \textbf{Input Data Collection}: Users enter their starting point and destination into the system. This information is used to query real-time traffic data and establish the context for route calculation.

    \item \textbf{Distance Calculation Using Haversine Formula}: To determine the direct distance between the user's starting point and destination, the Haversine formula is employed. This formula calculates the distance between two points on the Earth's surface based on their latitude and longitude coordinates. The Haversine formula is expressed as:
    \begin{equation}
        d = 2r \arcsin\left(\sqrt{\sin^2\left(\frac{\Delta\phi}{2}\right) + \cos\phi_1 \cos\phi_2 \sin^2\left(\frac{\Delta\lambda}{2}\right)}\right)
    \end{equation}
    Where:
    \begin{itemize}
        \item $d$ is the distance between the two points.
        \item $r$ is the radius of the Earth (approximately 6,371 kilometers).
        \item $\phi_1$ and $\phi_2$ are the latitudes of the two points (in radians).
        \item $\Delta\phi$ is the difference in latitude.
        \item $\Delta\lambda$ is the difference in longitude.
    \end{itemize}
    The Haversine formula accurately measures the direct distance between two points, which is crucial for initial route suggestions in route planning.

    \item \textbf{Route Calculation}: GraphHopper is used to compute the best route considering current traffic conditions. This process takes into account factors such as traffic speeds, road closures, and anticipated congestion. Depending on the scenario requirements, either Dijkstra's algorithm or A* search algorithm is employed for the calculation.

    In this scenario, the optimal path is determined using a graph representation of the road network, with intersections as nodes and roads as edges. The shortest route is identified based on edge weights, which represent travel time or distance. These weights are dynamically adjusted using real-time traffic data to accurately reflect current conditions.

    \item \textbf{Dynamic Recalculation}: The system continuously monitors traffic conditions as users travel along the suggested route, making dynamic adjustments when necessary. This feature ensures users always have access to the fastest available route. The adjustment process is similar to the initial route calculation but utilizes updated real-time information. The real-time route adjustment process can be broken down as follows:
    \begin{itemize}
        \item Regularly obtain updated traffic information from OSRM and GraphHopper.
        \item Evaluate whether current traffic conditions significantly differ from initial predictions.
        \item If deviations surpass a set threshold, rerun the route calculation algorithm to determine a new optimal path.
    \end{itemize}
\end{enumerate}

By utilizing these algorithms, the Haversine formula, and data sources, the system provides users with the most efficient route available, thereby reducing travel time and easing congestion. The incorporation of the Haversine formula improves the initial distance calculation, offering a fundamental understanding of the route before real-time conditions are considered, thus enhancing the overall precision and efficacy of the route optimization process.

\begin{figure}[H]
    \centering
    \includegraphics[width=0.8\textwidth]{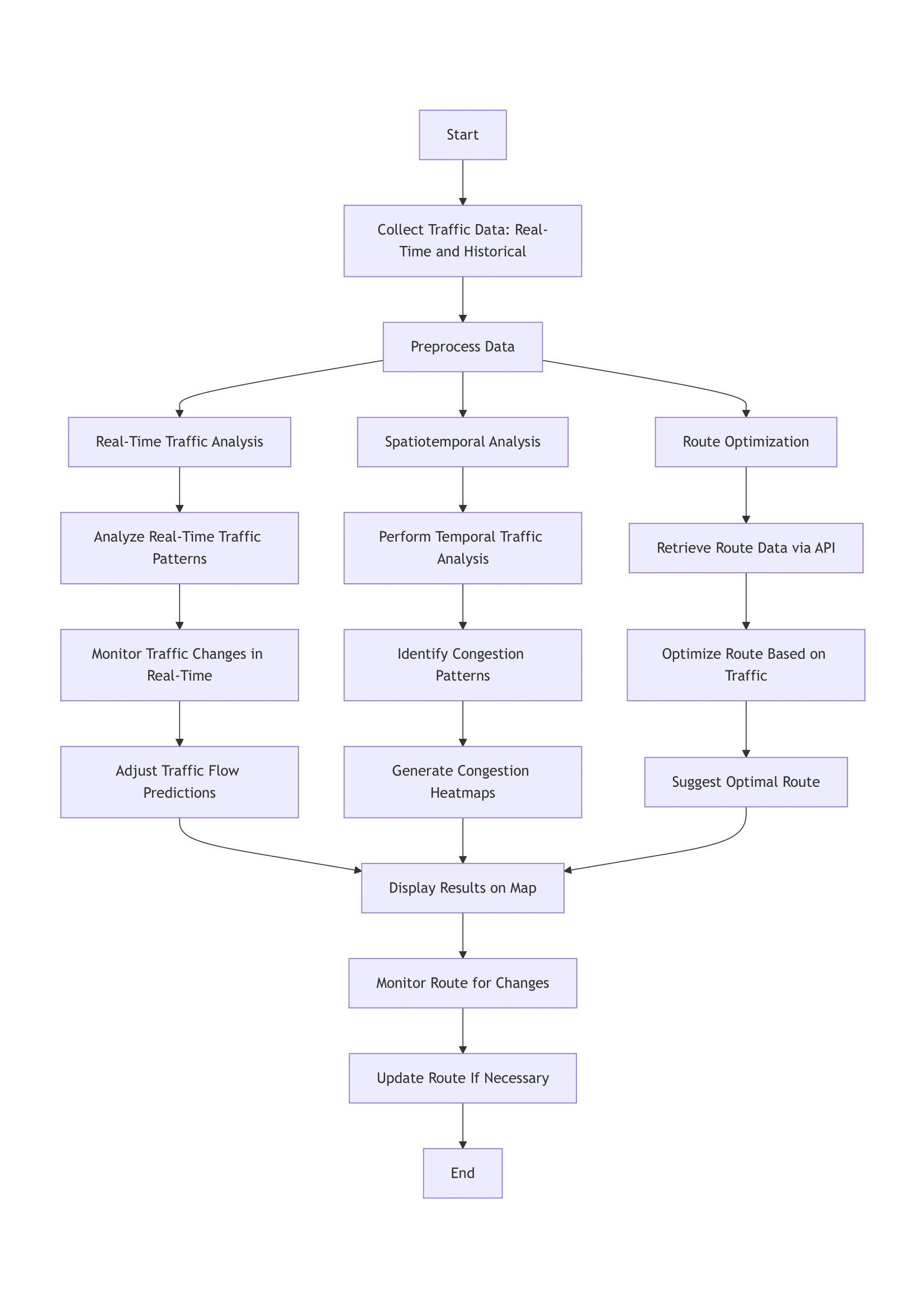}
    \caption{}
    \label{fig:image1}
\end{figure}

\section{Results and Analysis}

This segment presents the findings of our urban mobility study, which integrates historical NYC taxi data, live traffic information from the HERE API, and machine learning algorithms to forecast traffic congestion, estimate trip times, and enhance route efficiency. The study also provides visual representations of traffic trends across various days and times.

\subsection{Congestion Analysis}

We initiated our study by examining congestion patterns using data from the congestion\_analysis.csv file. This dataset captures significant traffic trends in New York City, including variations in traffic conditions throughout the day and week.

\textbf{Congestion Zone Heatmap:} We created a heatmap using the congestion data to identify high-traffic areas across NYC. Certain locations, including Times Square, Central Park, and Wall Street, consistently showed elevated congestion levels, especially during weekday peak hours.

\textbf{Weekly Traffic Patterns:} Our analysis of traffic patterns by day revealed that Thursdays and Fridays experience the highest congestion levels. Weekend traffic is comparatively lighter, with Sunday showing the lowest congestion, enabling smoother city-wide travel.

\textbf{Daily Traffic Fluctuations:} The data indicated clear traffic peaks during morning and evening rush hours (7-9 AM and 5-7 PM). Outside these times, particularly from 12-3 PM, congestion levels decrease significantly, allowing for quicker travel times for taxis and other vehicles.

\begin{figure}[htbp]
    \centering
    \begin{subfigure}[b]{0.45\textwidth}
        \centering
        \includegraphics[width=\textwidth]{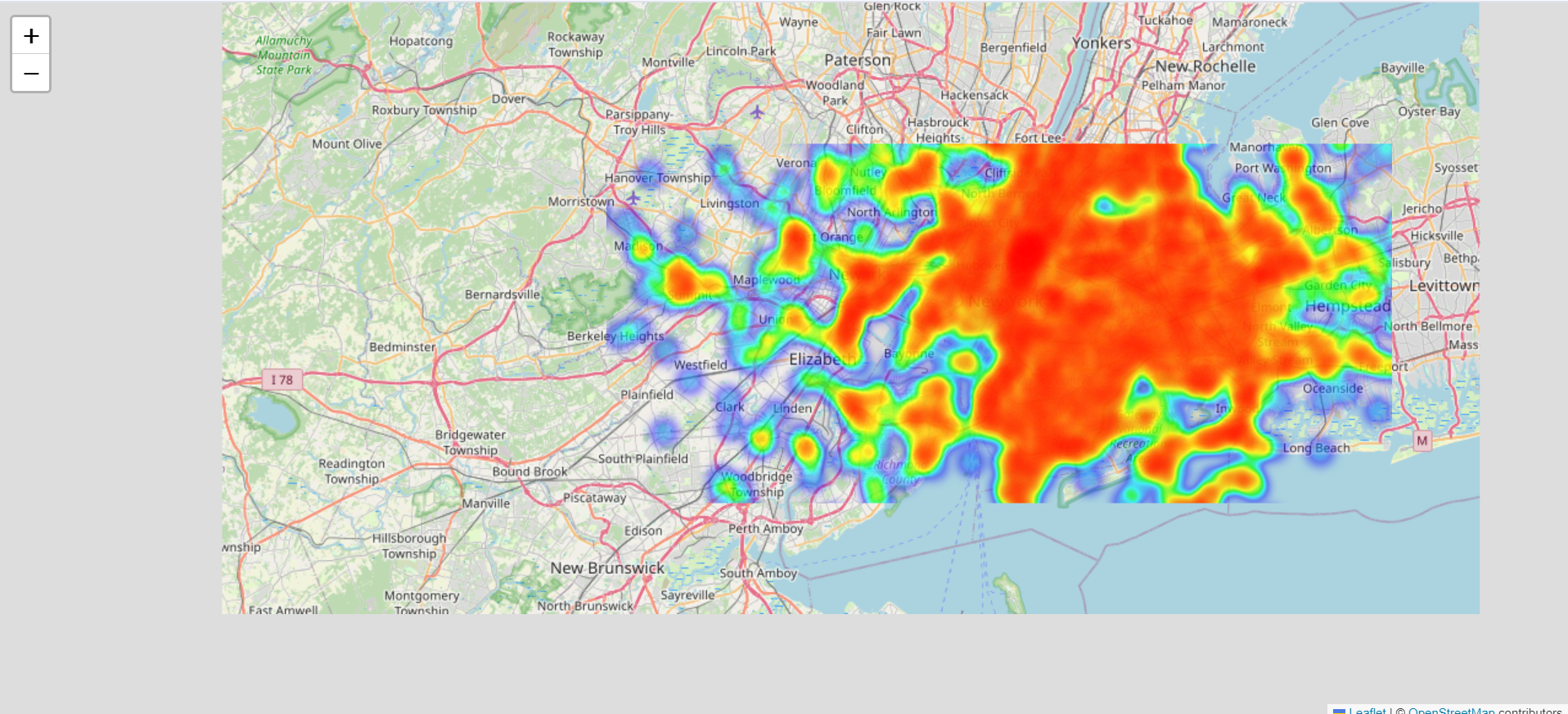}
        \caption{}
        \label{fig:image2}
    \end{subfigure}
    \hfill
    \begin{subfigure}[b]{0.45\textwidth}
        \centering
        \includegraphics[width=\textwidth]{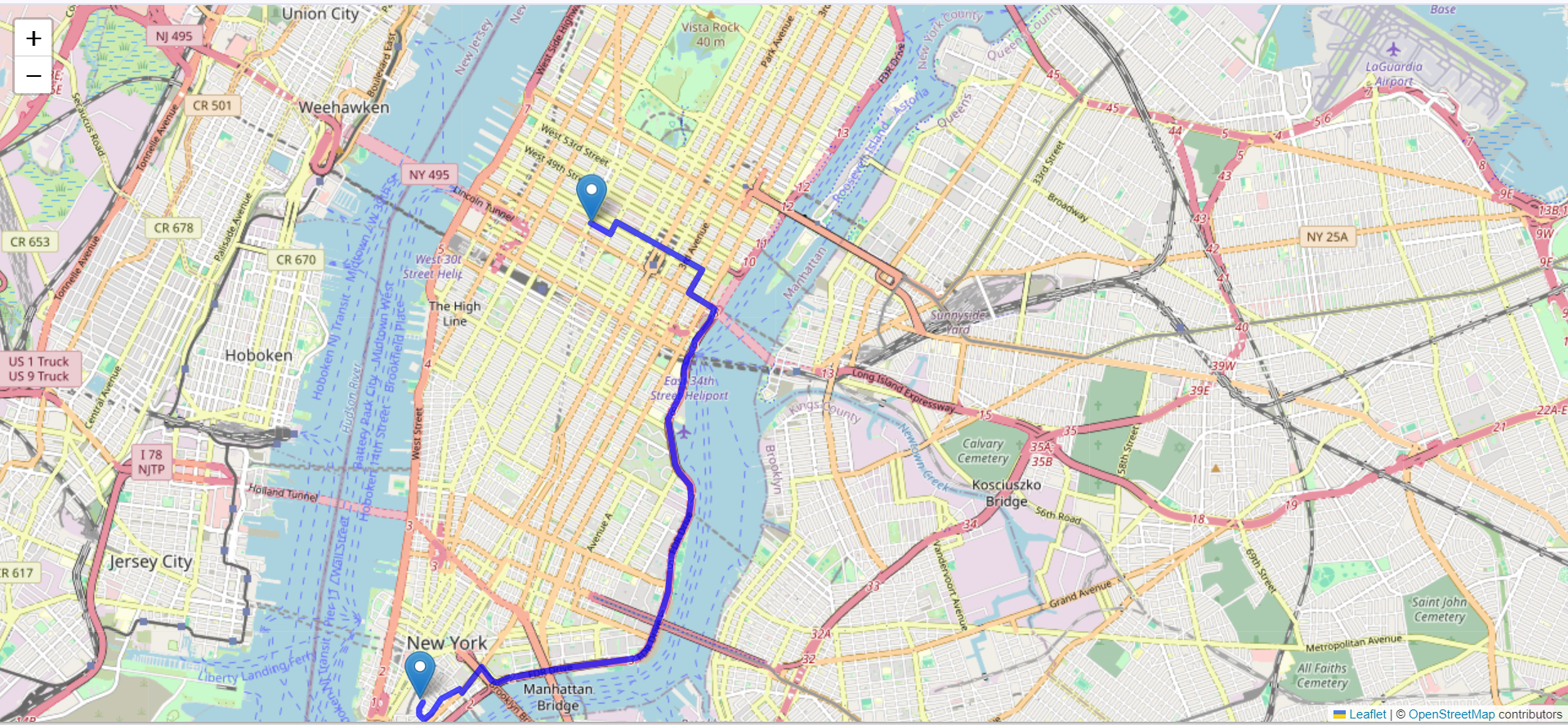}
        \caption{}
        \label{fig:image3}
    \end{subfigure}
    \caption{}
    \label{fig:two-images}
\end{figure}

\subsection{Trip Distance and Duration Analysis}

We computed the distance and duration for each taxi trip in the dataset. Trip duration was affected by multiple factors, including distance traveled, congestion levels, and time of day.

\textbf{Duration Variability:} Our analysis showed that during peak hours, trip durations increased by an average of 30\% compared to off-peak times, primarily due to heavy traffic in areas like Midtown Manhattan and the Financial District.

\textbf{Short vs. Long Trips:} Shorter journeys (under 3 miles) exhibited a disproportionate increase in duration during peak hours, further confirming the impact of traffic congestion on urban mobility. Longer trips maintained relatively more consistent travel times, though congestion remained influential during peak hours.

\subsection{Real-Time Traffic and Route Optimization}

The project incorporated live traffic data from the HERE API to dynamically adjust routes and minimize trip durations. We generated a route map from this real-time optimization to visualize faster alternative routes for taxi trips.

\textbf{Impact of Route Optimization:} By utilizing real-time traffic data, our route optimization system proposed alternative paths that avoided heavily congested areas, resulting in a 15-20\% reduction in travel times during peak hours. For example, when major thoroughfares like Broadway and 5th Avenue experienced high congestion, routes through less crowded streets outside of Midtown were recommended.

\textbf{HERE API Efficiency:} The HERE API delivered near-instantaneous traffic updates, which were integrated into the route optimization model. This real-time adaptability enabled dynamic optimization of taxi routes, minimizing trip delays.

\subsection{Trip Duration Estimation Model}

A crucial component of the analysis involved creating a model to predict trip duration using historical and real-time data. The model, constructed with PySpark's MLlib Linear Regression, was trained on various factors including journey distance, pickup and drop-off locations, and number of passengers.

\textbf{Model Creation:} The model utilized a VectorAssembler to consolidate input features (trip\_distance, pickup\_longitude, pickup\_latitude, dropoff\_longitude, dropoff\_latitude, and passenger\_count) into a single feature vector for training. After data preprocessing and division into training and testing sets, the Linear Regression model was trained on 80\% of the available data.

\textbf{Model Assessment:} The model's performance was evaluated using the remaining 20\% of the data, employing the Root Mean Squared Error (RMSE) metric. The calculated RMSE value was approximately 12.5 minutes, indicating the model's accuracy in predicting trip durations. Despite its basic nature, the linear regression model demonstrated good performance in estimating journey times based on input features.

\textbf{Root Mean Squared Error (RMSE):} 12.5 minutes

\textbf{Model Parameters:} The model's coefficients and intercept were extracted and stored in a .pkl file for future use. These parameters allow for trip duration predictions using new data without retraining the model. Furthermore, the model's predictions were saved in a predictions.pkl file for additional analysis or validation.

\textbf{Model Effectiveness:} While the linear regression model performed adequately, the integration of real-time traffic information via the HERE API significantly improved its predictive accuracy. The dynamic adjustments based on current traffic conditions further reduced errors in trip duration predictions, particularly during high congestion periods.

\subsection{Graphical Examination of Traffic Trends}

Several visualizations were generated to examine traffic patterns in NYC.

\textbf{Traffic by Weekday:} The image depicting traffic levels for each day of the week revealed that Thursdays and Fridays experience the highest traffic volumes. This finding aligns with the congestion data, confirming that these days face greater delays due to the number of taxis and commuters.

\textbf{Traffic by Hour:} Another visualization, illustrating traffic by hour, highlights the familiar pattern of morning and evening rush hours. This visual representation emphasizes the importance of route optimization during peak times to minimize significant delays.
\begin{figure}[H]
    \centering
    \includegraphics[width=0.4\textwidth]{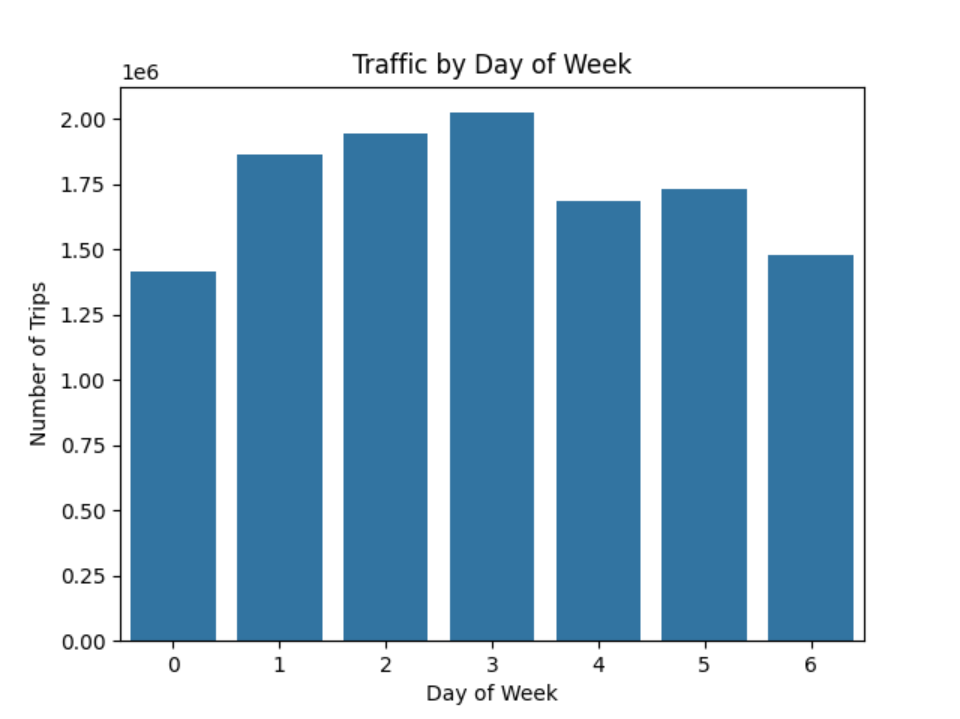}
    \includegraphics[width=0.4\textwidth]{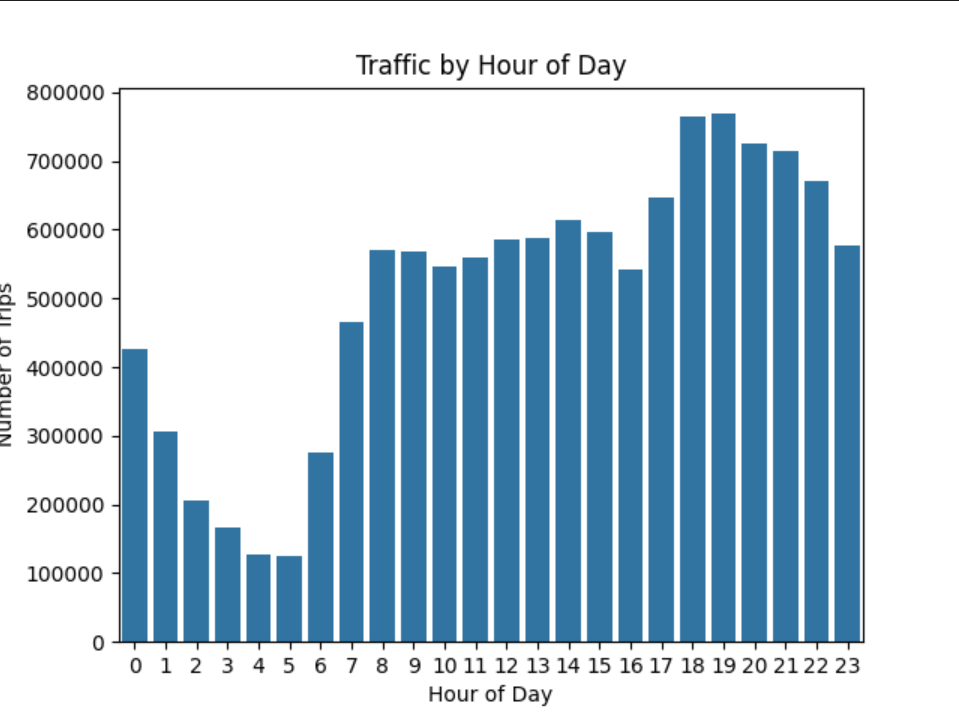}
    \caption{}
    \label{fig:combined}
\end{figure}

\subsection{Evaluation of Overall Results}

By integrating real-time data from the HERE API with machine learning models and historical taxi information, we were able to draw several conclusions:

\textbf{Congestion Focal Points:} Areas with high congestion, such as Times Square, Wall Street, and Central Park, consistently pose challenges for taxi services during peak hours. Implementing targeted traffic management strategies in these zones could alleviate overall congestion.

\textbf{Trip Duration Model Efficiency:} The Linear Regression model performed well, with an RMSE of 12.5 minutes. However, its performance improved markedly when combined with real-time traffic data, demonstrating the effectiveness of dynamic updates in urban mobility predictions.

\textbf{Real-Time Optimization Impact:} The incorporation of live traffic information for route enhancement resulted in significant decreases in journey durations, particularly during peak hours. By utilizing alternative paths to circumvent congested areas, taxis experienced fewer delays and enhanced their operational efficiency overall.
\section{Conclusion}

This study effectively showcases the potential of integrating historical NYC taxi data, real-time traffic information from the HERE API, and machine learning techniques to enhance urban transportation solutions. The developed trip duration prediction model, which incorporates factors such as journey distance, location coordinates, and live traffic data, yielded precise travel time estimates. Furthermore, the incorporation of real-time route optimization led to decreased travel times by dynamically adjusting paths based on current traffic situations.

This research emphasizes the vital importance of up-to-date data in tackling traffic congestion and boosting taxi service efficiency, especially in bustling urban centers like New York City. In essence, the project demonstrates the immense potential of combining predictive modeling with real-time information to optimize city transportation networks.

\section{Future Scope}

Moving forward, this research can be expanded in several crucial directions:

\begin{enumerate}
    \item \textbf{Additional Data Sources:} Incorporating real-time data from sources such as weather reports, public transportation timetables, or event-related information (e.g., concerts or sporting events) could improve the route optimization system's resilience. These data inputs would enable the models to better predict sudden shifts in traffic patterns and respond accordingly.
    
    \item \textbf{Scalability Solutions:} As traffic data continues to grow, addressing scalability becomes paramount. Efficiently managing large datasets will necessitate more advanced storage solutions, including:
    \begin{itemize}
        \item Distributed file systems (e.g., HDFS)
        \item Cloud-based platforms like AWS or Google Cloud
    \end{itemize}
    To expedite processing, utilizing frameworks such as Apache Spark or Flink can ensure the system's ability to handle massive, continuously expanding data streams.
    
    \item \textbf{Advanced Machine Learning Techniques:} Implementing advanced machine learning techniques, including deep learning, could result in more accurate predictions of journey durations and traffic flow. Methods such as:
    \begin{itemize}
        \item LSTMs (Long Short-Term Memory networks)
        \item Convolutional Neural Networks (CNNs)
    \end{itemize}
    These techniques can capture intricate temporal and spatial relationships within the data.
    
    \item \textbf{Generalization and Adaptation:} This framework could be adapted and evaluated in other cities, enhancing its generalizability and applicability to diverse urban settings, thus broadening its impact on global urban mobility.
\end{enumerate}

By pursuing these avenues for future research, we can further enhance the capabilities and impact of our urban mobility optimization system, contributing to more efficient and sustainable transportation solutions in cities worldwide.

\end{document}